%% file: main.tex
\documentclass{article}

\usepackage[preprint,nonatbib]{neurips_2021}

\usepackage{xcolor}         
\usepackage[utf8]{inputenc} 
\usepackage[T1]{fontenc}    
\usepackage{hyperref}       
\hypersetup{
	plainpages=false,
  colorlinks,
  linktocpage=true,
  linkcolor={red!50!black},
  citecolor={blue!50!black},
  urlcolor={blue!80!black}
} 

\usepackage{url}            
\usepackage{booktabs}       
\usepackage{amsfonts}       
\usepackage{nicefrac}       
\usepackage{microtype}      
\usepackage{enumitem}       
\newcommand{\subscript}[2]{$#1 _ #2$} 

\usepackage{multirow}
\usepackage{array,graphicx,subfigure}

\newcommand{\keypoint}[1]{\vspace{0.1cm}\noindent\textbf{#1}}

\newcommand{\ie}{\textit{i}.\textit{e}.}
\newcommand{\eg}{\textit{e}.\textit{g}.}
\newcommand{\etc}{\textit{etc}}

\newcommand{\sota}{state-of-the-art~}

\usepackage{amsmath,bm,amssymb,amsthm,amsfonts,amsopn}

\newtheorem{manualtheoreminner}{Theorem}
\newenvironment{manualtheorem}[1]{%
  \renewcommand\themanualtheoreminner{#1}%
  \manualtheoreminner
}{\endmanualtheoreminner}

\usepackage{mathtools}

\usepackage[capitalize,noabbrev]{cleveref}

\theoremstyle{plain}
\newtheorem{theorem}{Theorem}[section]
\newtheorem{proposition}[theorem]{Proposition}
\newtheorem{lemma}[theorem]{Lemma}

\theoremstyle{definition}

\newtheorem{assumption}[theorem]{Assumption}
\theoremstyle{remark}

\usepackage[textsize=tiny]{todonotes}

\newcommand{\dd}{\mathrm{d}}
\makeatletter
\newcommand*\bdot{\mathpalette\bdot@{.65}}
\newcommand*\bdot@[2]{\mathbin{\vcenter{\hbox{\hspace*{0.03em}\scalebox{#2}{$\m@th#1\bullet$}\hspace*{0.03em}}}}}
\makeatother

\title{How to Understand Masked Autoencoders}

\author{%
  Shuhao Cao\\
  Washington University in St. Louis\\
  \texttt{s.cao@wustl.edu} \\
   \And
   Peng Xu\thanks{Corresponding Author} \\
   University of Oxford \\
   \texttt{peng.xu@eng.ox.ac.uk} \\
   \AND
   David A. Clifton \\
   University of Oxford \\
   \texttt{david.clifton@eng.ox.ac.uk} \\
}

\begin{document}

\maketitle

\begin{abstract}
``Masked Autoencoders (MAE) Are Scalable Vision Learners'' revolutionizes the self-supervised learning method in that it not only achieves the state-of-the-art for image pre-training, but is also a milestone that bridges the gap between visual and linguistic masked autoencoding (BERT-style) pre-trainings.
However, to our knowledge, to date there are no theoretical perspectives to explain the powerful expressivity of MAE. 
In this paper, we, for the first time, 
propose a unified theoretical framework that provides a mathematical understanding for MAE.
Specifically, we explain the patch-based attention
approaches of MAE using an integral kernel under a non-overlapping domain decomposition setting.
To help the research community to further comprehend the main reasons of the great success of MAE, based on our framework, we pose five questions and answer them with mathematical rigor using insights from operator theory.
\end{abstract}

\section{Introduction}
\label{sec:introduction}

\begin{figure}[!t]
\centering
\includegraphics[width=0.7\columnwidth]{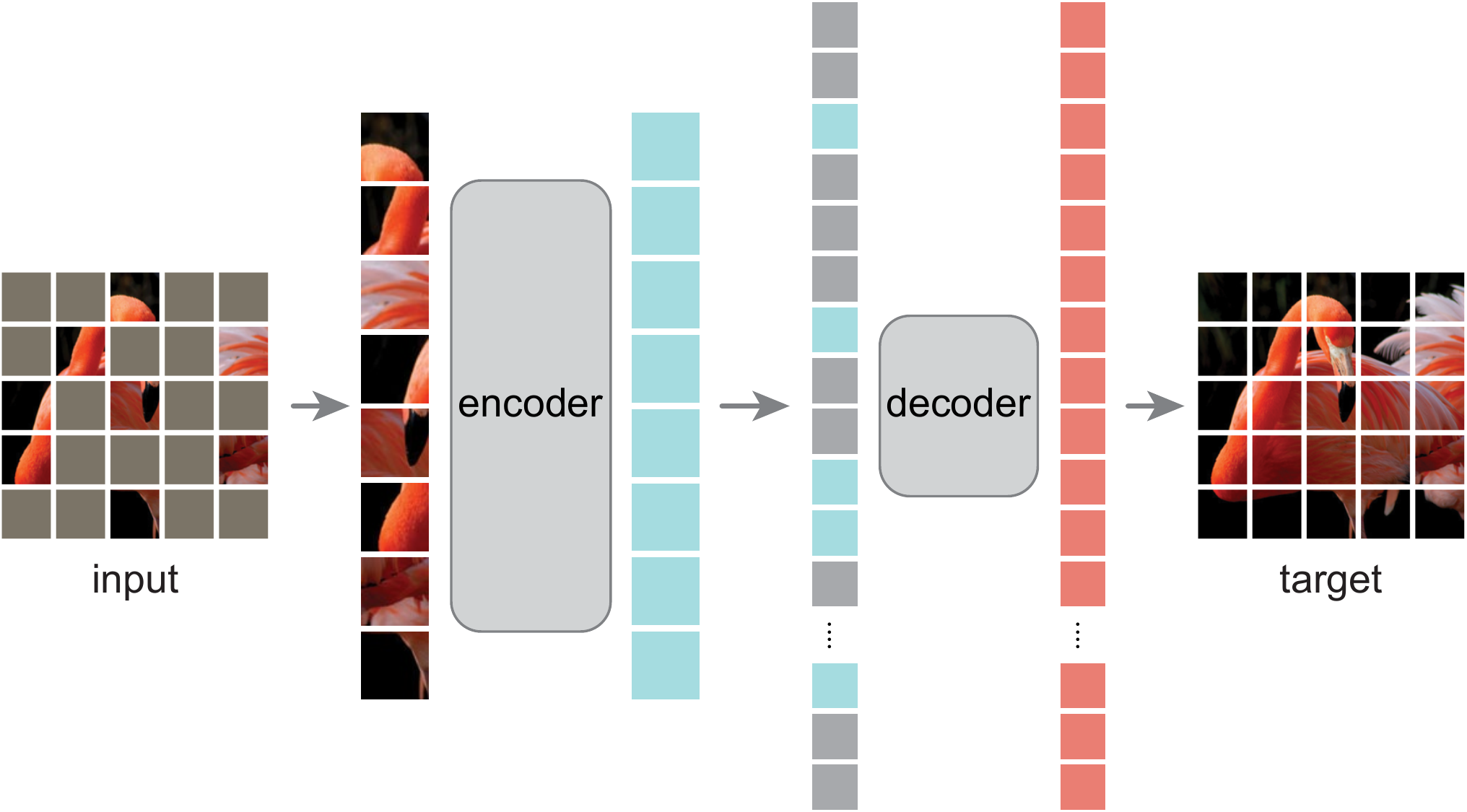}
\caption{Architecture of the Masked Autoencoder (MAE) in \cite{he2021masked}, with Vision Transformer (ViT) backbone. (Used under CC BY-NC 4.0 license from \url{https://github.com/facebookresearch/mae}.) Best viewed in color.}
\label{fig:mae}
\end{figure}

``Masked Autoencoders (MAE) Are Scalable Vision Learners''~\cite{he2021masked} (illustrated in Figure~\ref{fig:mae}) recently introduces a ground-breaking self-supervised paradigm for image pretraining.
This seminal method makes great contributions at least in the following respects:
\begin{enumerate}[leftmargin=1.5em,
                  label=(\arabic*)]
\item MAE achieves the \sota{} on self-supervised pretraining on ImageNet-1K dataset~\cite{deng2009imagenet}, 
outperforming the strong competitors (\eg, BEiT \cite{beit}) by a clear margin in a simpler and faster manner. 
Particularly, a \textit{vanilla}
Vision Transformer (ViT) \cite{DosovitskiyEtAl2020ViT} Huge backbone based MAE achieves the best accuracy ($87.8\%$) among methods that use only ImageNet-1K data. 
This inspiring phenomenon motivates the researchers to re-consider the ViT variants in self-supervised contexts. 
Moreover, MAE achieves good transfer performance
in downstream tasks beating the supervised pretraining
and shows promising scaling behavior. 

\item MAE is a generative learning method and beats the contrastive learning competitors (\eg, MoCo v3 \cite{chen2021empirical}) that have dominated vision self-pretraining in recent years.

\item MAE is a mile-stone that bridged the gap between the visual and linguistic masked autoencoding (BERT-style \cite{devlin2018bert}) pretrainings.
Actually, the previous work prior to MAE fails to apply the masked autoencoding pretrainings on the visual domain.
Thus, MAE paves a path that \textit{``Self-supervised learning in vision may now be embarking on a similar trajectory as in NLP''}~\cite{he2021masked}.
\end{enumerate}

The great success of MAE is interpreted by its authors as
\textit{``We hypothesize that this behavior occurs
by way of a rich hidden representation inside the MAE''}~\cite{he2021masked}.
However, to the best of our knowledge, to date, there are no theoretical viewpoints to explain the powerful expressivity of MAE. 
{Unfortunately, the theoretical analysis for the expressivity of ViT based models is so challenging that it is still under-studied.}



\textbf{Our Contributions} 
In this paper,
we,
for the first time, propose a unified theoretical
framework that provides a mathematical understanding for MAE. 
Particularly, we not only rethink MAE by regarding each image's embedding as a learned basis function in certain Hilbert spaces instead of a 2D pixel grid, but also explain the patch-based attention approaches of MAE from
the operator theoretic perspective of an integral kernel under a non-overlapping domain decomposition setting.

To help the researchers to further grasp the main reasons for the great success of MAE, based on our proposed unified theoretical framework, we contribute five questions, and for the first time answer them partially by insights from rigorous mathematical theory: 

\textbf{Q1:} 
\textit{How is the representation space of MAE formed, optimized, and propagated through layers?}
\\
\textbf{A:}  We illustrate that the attention mechanism in MAE is equivalent to a learnable integral kernel transform, and its representation power is dynamically updated by the Barron space with the positional embeddings {that work as the coordinates in a high dimensional feature space}. See Section \ref{sec:self-attention}. 

\textbf{Q2:} \textit{Why and how does patchifying contribute to MAE?} \\
\textbf{A:} We prove that the random patch selecting of MAE preserves the information of the original image, while reduces the computing costs, under common assumptions on the low-rank nature of images. {This paves a theoretical foundation for the patch-based neural networks/models including but not limited to MAE or ViT.} See Section \ref{sec:patch}.

\textbf{Q3:} 
\textit{Why are the internal representations in the lower and higher layers of MAE not significantly different?} \\
\textbf{A:} We have provided a new theoretical justification for the first time by proving the stability of the internal presentations. The great success of MAE benefits from a main reason that the scaled dot-product attention in the ViT-backbone provides stable representations during the cross-layer propagation. Furthermore, we view the skip-connection for the attention mechanism from a completely new perspective: representing an approximated solution explicitly to a Tikhonov-regularized Fredholm integral equation. See Section \ref{sec:vit-stability}.

\textbf{Q4:} \textit{Is decoder unimportant for MAE?} \\
\textbf{A:} No. 
We argue that the decoder is vital to helping encoder to build better representations, even if decoder is discarded after pretraining. Due to the bigger patch dimension in the MAE decoder, it allows the representation space in the encoder enriched much more often by functions from the Barron space to learn a better basis.
See Section~\ref{sec:MAE-decoder}. 

\textbf{Q5:} \textit{Does MAE reconstruct each masked patch merely inferred from its adjacent neighbor patches?} \\ 
\textbf{A:} No. We prove that
the latent representations of the masked patches
are interpolated globally based on an inter-patch
topology that is learned by the attention mechanism. See Section~\ref{sec:MAE-decoder}. 

Overall, our proposed unified theoretical framework provides a mathematical understanding for MAE and can be used to understand 
the intrinsic traits of the extensive patch and self-attention based models, not limited to MAE or ViT.

The rest of this paper is organized as follows: Section~\ref{sec:related-work} summarizes related work. 
Based on our proposed theoretical framework,
some mathematical understandings for the encoder and decoder of MAE are presented in Section~\ref{sec:MAE-encoder} through Section~\ref{sec:MAE-decoder}.
We draw some conclusions in Section~\ref{sec:conclusion}. 

\section{Related Work}
\label{sec:related-work}


\keypoint{Vision Transformer (ViT)} \cite{DosovitskiyEtAl2020ViT} is a strong image-oriented network based on a standard Transformer \cite{vaswani2017attention} encoder.
It has an image-specific input pipeline where the input image needs to be split into fixed-size patches.
After going through the linearly embed layer and adding with position embeddings, all the patch-wise sequences will be encoded by a standard Transformer encoder.
ViT and its variants have been widely used in various computer vision tasks (\eg, recognition \cite{touvron2021training}, detection \cite{beal2020toward}, segmentation \cite{liu2021swin}), and meanwhile work well for both supervised \cite{touvron2021training} and self-supervised \cite{chen2021empirical,caron2021emerging} visual learning.
Recently, some pioneering works provide further understanding for ViT, \eg, its internal representation robustness \cite{PaulChen2022Vision}, the continuous behavior of its latent representation propagation \cite{RaghuUnterthinerEtAl2021ViT}.
However, the theoretical analysis for the expressivity of ViT based models is so challenging that it is still under-studied.


\keypoint{Masked Autoencoder (MAE)} \cite{he2021masked} is essentially a denoising autoencoder \cite{vincent2008extracting}, which has a straightforward motivation that randomly mask patches of the input image and reconstruct the missing pixels.
MAE works based on two key designs:
(i) asymmetric encoder-decoder architecture where the encoder takes in only the visible patches and the lightweight decoder reconstructs the target image,
(ii) high masking ratio (\eg, $75\%$) for the input image yields a nontrivial and meaningful self-supervisory task.
Its great success is attributed to 
\textit{``a rich hidden representation inside the MAE''}~\cite{he2021masked}.
However, to the best of our knowledge, to date, there are no theoretical viewpoints to explain the powerful expressivity of MAE.

\keypoint{Mathematical Theory Related to Attention}
Self-attention \cite{vaswani2017attention} is essentially processing each input as a fully-connected graph \cite{xu2021multigraph}.
Therefore, as aforementioned, we start from a more general perspective of topological spaces \cite{carlsson2009topology} to rethink MAE, by regarding
each image as a graph connecting patches instead of 2D pixel grid.
Meanwhile, we study the patch-based attention approaches of MAE through the operator theory for the Fredholm integral equations \cite{AtkinsonHan2009Numerical}, to formulate the dot-product attention matrix as an integral kernel \cite{HalmosSunder2012Bounded}, \ie, a learned Green's function to reconstruct solutions to the partial differential equations \eg. \cite{GinSheaEtAl2021DeepGreen}. The skip-connection can be then explained through two perspectives, first as the term corresponding to the Tikhonov regularization \cite{Groetsch1984theory,Weese1992reliable}, or 
Neural Ordinary Differential Equations (ODE) \cite{chen2018neural}. On the other hand, the patchification is connected with the Domain Decomposition Methods (DDM) \cite{ToselliWidlund2004Domain,ZhangBurgerEtAl2010Bregmanized}.















\input{2_vit.tex}

\input{3_mae.tex}

\section{Conclusion}
\label{sec:conclusion}

To the best of our knowledge, to date, there are no theoretical viewpoints to explain the powerful expressivity of MAE.
In this paper, we, for the first time, 
propose a unified theoretical framework that provides a mathematical understanding for MAE.
Particularly, we explain the patch-based attention
approaches of MAE from a perspective of an integral kernel under a non-overlapping domain decomposition setting.
To help the researchers to further grasp the main reasons for the great success of MAE, our mathematical proof contributes the following major conclusions: 

\textbf{(1)}  The attention mechanism in MAE is a learnable integral kernel transform, and its representation power is dynamically updated by the Barron space with the positional embeddings {that work as the coordinates in a high dimensional feature space}.

\textbf{(2)} The random patch selecting of MAE preserves the information of the original image, while reduces the computing costs, under common assumptions on the low-rank nature of images. {This paves a theoretical foundation for the patch-based neural networks/models including but not limited to MAE or ViT.} 

\textbf{(3)} The great success of MAE benefits from a main reason that the scaled dot-product attention built-in ViT provides the stable representations during the cross-layer propagation. 

\textbf{(4)}
In MAE, the decoder is vital to helping the encoder to build better representations, while decoder is discarded after pretraining.

\textbf{(5)} 
The latent representations of the masked patches
are interpolated globally based on an inter-patch
topology that is learned by the attention mechanism.


Furthermore, our proposed theoretical framework can be used to understand the intrinsic traits of not only the ViT-based models but also even the extensive networks/models made by patch and self-attention.

{
\small

}



\input{4_appendix.tex}



\end{document}

%% file: 2_vit.tex

\section{Patch is All We Need?}
\label{sec:patch}

Patchifying has become a standard practice in Transformer-based CV models since ViT \cite{DosovitskiyEtAl2020ViT}, see also \cite{HanEtAl2021transformer,LiuLinEtAl2021Swin}. In this section, we try to answer the question ``Why and how does patchifying contribute to MAE?'' from a perspective of the domain decomposition method to solve an integral equation. 

We first consider the full grid $\Omega \simeq I\otimes I $ with $I = [0, 1/N, \dots, (N-1)/N]$,
then 
the patchification of $\Omega$ is essentially a non-overlapping domain decomposition method (DDM) \cite{ChanMathew1994Domain}. DDM is commonly used in solving integral/partial differential equations bearing similar forms with \eqref{eq:attention-kernel}, {\eg,} see \cite{ToselliWidlund2004Domain}, as well as image denoising problems \cite{ZhangBurgerEtAl2010Bregmanized}. 

DDM practices ``divide and conquer'', which is a general methodology suitable for lots of science and engineering disciplines: the original domain of interest is decomposed into much smaller subdomains. Each subproblem associated with subdomains can be more efficiently solved by the same algorithm, especially when the algorithm has a super-linear dependence on the grid size. The attention mechanism computational complexity and storage requirement have a quadratic dependence on $n$, and this translates a quartic dependence on the coarse grid size $p$ (how many patches along one axis) as $n=p^2$. Without patchification, we can see the operation in \eqref{eq:attention-dot-product} is of $\mathcal{O}(N^4)$ that renders the algorithm unattainable. 

In this setup, $\Omega$ is first partitioned to equal-sized non-overlapping subdomains: $\Omega = \cup_{i=1}^n \Omega_i$, $\operatorname{int}(\Omega_i)\cap \operatorname{int}(\Omega_j) =\emptyset$ as well as $\operatorname{int}(\Omega_i)\simeq \operatorname{int}(\Omega_j)$  if $i\neq j$. For the simplicity of the presentation, we consider a single-channel image and define the space of bounded variation (BV) on $\Omega$ as $BV(\Omega)$, and for any unpatchified image $\mathbf{u}\in BV(\Omega)\subset \mathbb{R}^{N\times N}$
\begin{equation}
\label{eq:norm-total-variation}
|\mathbf{u}|_{BV(\Omega)} := 
\sum_{x_i \in \Omega}\left\{\sum_{x_j \in N_{x_i}} w_{ij} (\mathbf{u}(x_i)-\mathbf{u}(x_j))^{2}\right\}^{\frac{1}{2}},
\end{equation} 
where $N_{x_i}$ denotes the grid points that are connected with $x_i$ through an undirected edge $x_i x_j$, and $w_{ij}$ is some positive weights measuring the interaction strength.
When being viewed in the continuum such that $\mathbf{u}$ is a discrete sampling of the function $u: \mathbb{R}^2\to \mathbb{R}^+$, this norm approximates $|\nabla u|_{L^1}$ and measures the smoothness of an image, and is widely used in graph/image recovery \eg, see \cite{BergerHannakEtAl2017Graph}.

After the patchification 
we have $\mathbf{u} = \sum_{i=1}^{n^2} E_i \mathbf{u}_i$, 
where $E_i:\Omega_i \to \Omega$ is an extension operator that maps the patch pixel values on $\Omega_i$ to $\Omega$ by zero-padding. It is straightforward to see that the representation consist of the concatenated patch matrices, and
is in $\Pi^{n^2}_{i=1} BV(\Omega_i)$. Here we denote $\mathcal{Y}:=\oplus_{i=1}^{n^2} E_i\, BV(\Omega_i)$. 
By \eqref{eq:norm-total-variation}, the original BV space is only a subspace of the decomposed product space, \ie, $BV(\Omega)\subset \mathcal{Y}$, while the reverse is not true. The reason is that by patchification, the underlying function spaces have completely lost the inter-patch topological relations: 
\eg, how the inter-patch pixel intensities change?
There is also another analytical set of questions to answer: how to choose the size of the patch? If bigger patches are used, it then requires a bigger embedding dimension, thus harder to formulate the bases for the representation space. On the other hand, if smaller patches are used, because of the aforementioned quartic dependence on the grid size, the inter-patch topology is much harder and more expensive to learn.


{In the following proposition, we shall show that, if an image has a low rank structure, then there exists a semi-randomly chosen set of patches that can represent the original image. This selection has a very high mask ratio, that is, the representation using a small fraction e.g., $p$ patches out of all the patches. Meanwhile, we assume that there exists an embedding that is able to represent the original image, in the sense that the difference between the reconstruction and the original image is small measured under $|\cdot|_{BV(\Omega)}$.}

Following the standard practices \cite{DosovitskiyEtAl2020ViT, touvron2021training}, we consider the ViT-base case in MAE where the dimension of the feature embedding matches the original dimension of the pixel counts in the any given patch: $\mathbf{u}\in \mathbb{R}^{N\times N}\simeq \mathbb{R}^{n^2\times d}$, \ie, for each patch's embedding in the MAE encoder, there exists a continuous embedding that maps $\mathbb{R}^{n^2\times d}\hookrightarrow \mathbb{R}^{N\times N}$, from the patch embeddings to images. 

\begin{proposition}[Existence of a near optimal patch embedding]
\label{theorem:low-rank}
For any $\mathbf{u}\in BV(\Omega)\subset \mathbb{R}^{N\times N}$, and an equal-sized $n\times n$ patchification of $\mathbf{u}$, let the patch grid size $N_c := N/n$, assuming there exists a rank $r$ approximation to $\mathbf{u}$ with $r<N_c$ and error $\epsilon$, and a unique patch is randomly chosen from each row such that the final selection's columns set is a permutation of $\{1,\cdots, n^2\}$, then there exists an embedding $\mathbf{y}\in \mathbb{R}^{p\times d}$ such that 
\begin{equation}
\|\mathbf{u} - R (\mathbf{y}) \|_{BV(\Omega)} <  C(r, N_c, p) \epsilon,
\end{equation}
where $R$ is a rank preserving reconstruction operator $R: \mathbb{R}^{p\times d} \to \mathbb{R}^{N\times N}$.
\end{proposition}







\section{Attention in MAE: a Kernel Perspective}
\label{sec:MAE-encoder}
In this section, we reexamine the attention block present in both encoder and decoder layers in MAE from multiple perspectives. The first and foremost understanding comes from the operator theory for the integral equations. By formulating the scaled dot-product attention as a nonlinear integral transform, learning the latent representation is equivalent to learning a set of basis in a Hilbert space. 
Moreover, upon adopting the kernel interpretation of the attention mechanism, a single patch is characterized by not only its own learned embedding, but also how it interacts with all other patches (reproducing property of a kernel).



\subsection{Self-attention: A Nonlinear Integral Kernel Transform}
\label{sec:self-attention}

For MAE \cite{he2021masked}, each input image patch is projected as a 1D token embedding. 
Following the practice in \cite{he2021masked},
we also omit the shared class token added to the embedding that can be removed from the pretraining of MAE. 
Given an attention block in an encoder or a decoder layer of MAE, the input and output are defined respectively as 
$\mathbf{y}_{\text{in}}:=\mathbf{y}, \mathbf{y}_{\text{out}} \in \mathbb{R}^{p\times d}$. 
When an image has not been masked, it has $p=n^2$ patches, where $n$ is a positive integer. The embedding dimension is $d$. 
{
In $\mathbf{y}$, a positional embedding $\mathcal{X}:=\{x_i\}_{i=1}^p\subset \mathbb{R}^{1\times d}$, such that $x_i$ is associated with the $i$-th patch, is added. Each $x_i$ can be viewed as a coordinate in high dimension.}
We note that the patch ordering map $i\mapsto x_i$ is injective, since the first component of $x_i$ is the polar coordinate in 2D, and as such the relative position of each patch can be recovered from $x_i$. 
For simplicity, the analysis done in this section applies to a single head.

\begin{figure*}[!t]
\centering
\includegraphics[width=0.99\textwidth]{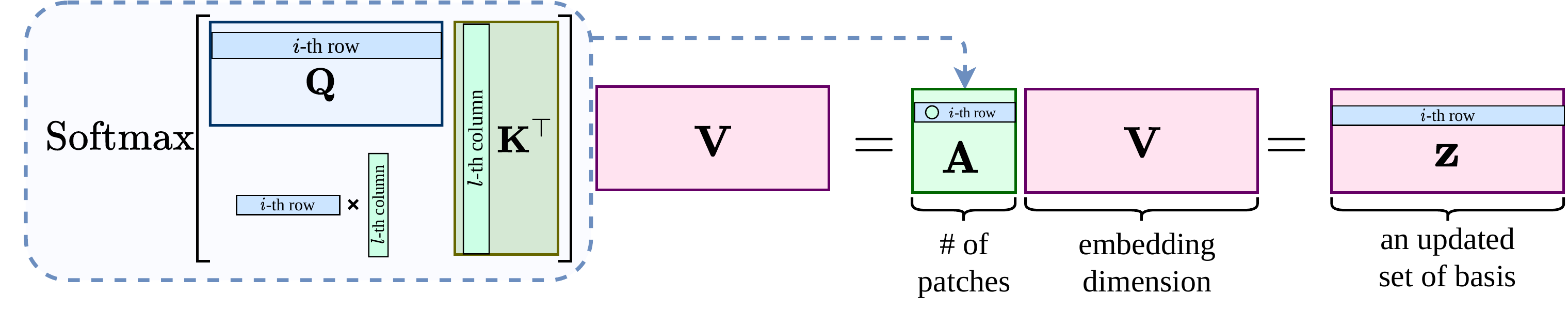}
\caption{The $i$-th patch embedding from the scaled dot-product attention is a convex combination 
of the attention weights,
which encodes the interactions among all patch embeddings. Best viewed in color.}
\label{fig:attention}
\end{figure*}

Here we briefly review the self-attention from \cite{vaswani2017attention}. The query $\mathbf{Q}$, key $\mathbf{K}$, value $\mathbf{V}$ are generated by three learnable projection matrices $\mathbf{W}^Q, \mathbf{W}^K, \mathbf{W}^V \in \mathbb{R}^{d\times d}$: $\mathbf{Q} = \mathbf{y} \mathbf{W}^Q$, $\mathbf{K} = \mathbf{y} \mathbf{W}^K$, $\mathbf{V} = \mathbf{y}\mathbf{W}^V$.
The scaled dot-product attention is to obtain $\mathbf{z} \in \mathbb{R}^{p\times d}$:
\begin{equation}
\label{eq:attention}
    \mathbf{z} = \text{Attn}_s (\mathbf{y}) := \operatorname{Softmax}\left(\mathbf{Q} \mathbf{K}^\top / \sqrt{d} \right) \mathbf{V}.
\end{equation}
Then the 
softmax attention $\operatorname{Attn}(\cdot)$ with a global receptive field works as the following nonlinear mapping:
\begin{equation}
\begin{aligned}
\operatorname{Attn}: & \; \mathbb{R}^{p\times d}\to \mathbb{R}^{p\times d}, 
\\
& \; \mathbf{y} \mapsto \operatorname{LN}
\left(\mathbf{y}+\mathbf{z} + \operatorname{FFN}\Big(\operatorname{LN}(\mathbf{y}+\mathbf{z}) ) \Big)\right),
\end{aligned}
\end{equation}
where $\operatorname{LN}(\cdot)$ denotes the Layer Normalization (LN) \cite{ba2016layer} that essentially is a learnable column scaling with a shift, and $\operatorname{FFN}(\cdot)$ is a standard two-layer feedforward neural network (FFN) applied to the embedding of each patch.

Upon closer inspection of the scaled dot-product attention~\eqref{eq:attention},
for $\mathbf{z}$, the $j$-th element of its $i$-th row $\mathbf{z}_i$ is obtained by
\begin{equation}
(\mathbf{z}_i)_j = \mathbf{A}_{i \bdot} \cdot \mathbf{v}^j,
\end{equation}
and the $i$-th row $\mathbf{A}_{i \bdot}$ of the attention matrix $\mathbf{A}$ is
\begin{equation}
\label{eq:attention-dot-product}
\mathbf{A}_{i \bdot} =  \frac{e^{(\mathbf{Q} \mathbf{K}^{\top}/\sqrt{d})_{i \bdot}}}{
\sum_{j=1}^p e^{(\mathbf{Q} \mathbf{K}^{\top} /\sqrt{d})_{i j}}} 
=: \operatorname{Softmax} (\mathbf{q}_i \mathbf{K}^{\top}/\sqrt{d}), 
\end{equation}
\ie, using the diagram in Figure \ref{fig:attention}, we have
\begin{equation}
  \mathbf{z}_i = 
\mathbf{A}_{i \bdot} \begin{pmatrix}
\rule[.4ex]{2em}{0.2pt} & \mathbf{v}_1  &\rule[.4ex]{2em}{0.2pt}
\\[-4pt]
\rule[.8ex]{2em}{0.3pt} & \vdots & \rule[.8ex]{2em}{0.3pt}
\\[-1pt]
\rule[.5ex]{2em}{0.2pt}&  \mathbf{v}_n &  \rule[.5ex]{2em}{0.2pt}
\end{pmatrix}
= \sum_{j=1}^p \mathbf{A}_{ij} \mathbf{v}_j.
\end{equation}
Due to the softmax, 
{$\mathbf{A}_{i \bdot}$ contains the coefficients for the convex combination of the vector representations $\{\mathbf{v}_j\}_{j=1}^p$. 
This basis set $\{\mathbf{v}_j\}_{j=1}^p$ form the $\mathbf{V}$'s row space, and it further forms each row of the output $\mathbf{z}$ by multiplying with $\mathbf{A}$.} 
As a result, the scaled dot-product attention has the following basis expansion form:
\begin{equation}
\label{eq:attention-expansion}
  \mathbf{z}_i := \sum_{j=1}^p 
A(\mathbf{q}_i, \mathbf{k}_j) \, \mathbf{v}_i,
\end{equation}
where $\mathbf{u}_i$ denotes the $i$-th row of certain latent representation $\mathbf{u}\in \{\mathbf{z}, \mathbf{Q}, \mathbf{K}, \mathbf{V}\}$. {The key term in \eqref{eq:attention-expansion}, $A(\cdot, \cdot)$, denotes the attention kernel, which maps each patch's embedding represented by the rows of $\mathbf{Q}, \mathbf{K}$ to a measure of how they interact.} From~\eqref{eq:attention-expansion}, we shall see that the representation space for an encoder layer in MAE is spanned by the row space of $\mathbf{V}$, and is being nonlinearly updated layer-wise. This means, the embedding for each patch serves as a basis to form the representation space {for the current attention block}, whereas the
row-wise attention weights 
are the Barycentric coordinates for a simplex in $\mathbb{R}^{1\times d}$.

Here we further assume that there exist a set of feature maps for query, key, and value \eg, see \cite{ChoromanskiEtAl2020performer}. For $\mathbf{z}$, the feature map $z\in \mathcal{V}$ that maps $\mathbb{R}^{1\times d} \to BV(\Omega_i)$, \ie, $\mathbf{z}_i = z(x_i)$, we can then define an asymmetric kernel function $\tilde{\kappa}(\cdot, \cdot): \mathbb{R}^{1\times d} \times \mathbb{R}^{1\times d}\to \mathbb{R}$,
\begin{equation}
\begin{aligned}
A (\mathbf{q}_i, \mathbf{k}_j)
&= \alpha^{-1}(x_i) \langle \mathbf{q}_i, \mathbf{k}_j \rangle
\\
&= \alpha^{-1}(x_i) \langle q(x_i), 
k(x_j) \rangle
\\
&=: \alpha^{-1}(x_i)\, \tilde{\kappa}(x_i, x_j),
\end{aligned}
\end{equation}
where $\tilde{\alpha}(x_i) := \sum_{l=1}^d (e^{\mathbf{q}_i \mathbf{K}^{\top}/\sqrt{d}})_l$. Now the discrete kernel $A(\cdot,\cdot)$ with vectorial input is rewritten to an integral kernel $\tilde{\kappa}(\cdot,\cdot)$ whose inputs are positions.
As a result, using the formulation above, we can express the scaled dot-product attention approximately as a nonlinear integral transform:
\begin{equation}
\label{eq:attention-kernel}
\begin{aligned}
z(x) & = \alpha^{-1}(x) \sum_{x'\in \mathcal{X}} \big(q(x) \cdot k(x')\big) v(x') \, \delta_{x'}
\\
&\approx \alpha^{-1}(x)\int_{\omega} \tilde{\kappa}(x, x') v(x') \,d\mu(x'),
\end{aligned}
\end{equation}
where $\delta_{x}$ is the Dirac measure associated with the position at $x$. For the second equation, with a slight abuse of the order presentation, we assume that there exists $\tilde{\kappa}(\cdot,\cdot)$ and a Borel measure $\mu(\cdot)$ such that $ \tilde{\kappa}(\cdot,\cdot)$ has a reproducing property under the integral by $d\mu(\cdot)$, which shall be elaborated in the next paragraph. {Here in this integral, the image domain $\omega$ is approximated by a patchified grid $\mathcal{X} \simeq (0, \frac{1}{n}\cdots, \frac{n-1}{n})\otimes (0, \frac{1}{n}\cdots, \frac{n-1}{n})$.} Returning to the perspective of ``embedding $\approx$ basis'' for the representation space, the backpropagation from the output $z(\cdot)$ to update weights to obtain a new $v(\cdot)$ during pretraining can be interpreted as an iterative procedure to solve the Fredholm integral equation of the {first kind} in \eqref{eq:attention-kernel}.

\paragraph{How to Obtain a Translation Invariant Kernel.} 
\label{paragraph:rkhs}
We note that the formulation above in \eqref{eq:attention-kernel} is closely related to Reproducing Kernel Hilbert Space (RKHS) \cite{berlinet2011reproducing}. First, the dot-product in \eqref{eq:attention-dot-product} is re-defined to be
\begin{equation}
\label{eq:kernel-gaussian}
\mathbf{A}_{i \bdot} := \operatorname{Softmax} \Big(-\gamma (\mathbf{q}_i-\mathbf{k}_i) (\mathbf{Q}-\mathbf{K})^{\top}\Big).
\end{equation}
Then, a pre-layer normalization scheme in \cite{baevski2018adaptive, child2019generating, wang2019learning, xiong2020layer} or pre-inner product normalization in \cite{HenryDachapallyEtAl2020Query,Cao2021transformer} and can be a cheap procedure to re-center the kernel.
Rewritting the original dot-product for $\mathbf{q},\mathbf{k}\in \mathbb{R}^{1\times d}$ as follows 
\begin{equation}
\label{eq:kernel-shift}
\mathbf{q} \mathbf{k}^{\top} = 
\frac{1}{2} \mathbf{q} \mathbf{q}^{\top} + \frac{1}{2} \mathbf{k} \mathbf{k}^{\top} - \frac{1}{2} (\mathbf{q}-\mathbf{k}) (\mathbf{q}-\mathbf{k})^{\top}.
\end{equation} 
Thus, a layer normalization for each row in $\mathbf{Q}$ and $\mathbf{K}$ before $\mathbf{Q}\mathbf{K}^{\top}$ {results the two moment terms above, $\mathbf{q} \mathbf{q}^{\top}$ and $\mathbf{k} \mathbf{k}^{\top}$, in \eqref{eq:kernel-shift} being relatively small versus the third term. Consequently, using merely the third term in \eqref{eq:kernel-shift} to form \eqref{eq:kernel-gaussian} suffices to provided a good approximation to the scaled dot-product attention.}
As an important consequence, the kernel becomes symmetric, and translation-invariant, which is arguably one of the nicest traits of CNN kernels. As a result, the normalized attention kernel $\tilde{\kappa}(x, x')$ based on \eqref{eq:kernel-gaussian} is
\begin{equation}
\label{eq:attention-normalized}
(\mathbf{A})_{ij}= \gamma \langle \mathbf{q}_i - \mathbf{k}_i, \mathbf{q}_j - \mathbf{k}_j \rangle =: \kappa(x_i, x_j).
\end{equation}
Moreover, $\kappa(x,x')$ can be written as $K(x-x')$ for $K: \mathbb{R}^{1\times d}\to \mathbb{R}$ being the 1-dimensional Gaussian radial basis function (RBF) kernel, $K(\theta) = e^{-\gamma \|\theta\|^2}$.
By the positivity and the symmetrcity, $\kappa(\cdot,\cdot)$ becomes a reproducing kernel. We can define the following nonlinear integral operator $\mathcal{K}$:
\begin{equation}
    \mathcal{K}(v)(x) := 
\int_{\omega} \kappa(x, x') v(x') \,d\mu(x').
\end{equation}
{In the traditional settings such that $\kappa(x, x')$ is not explicitly dependent on $v(\cdot)$, the vector version of the Mercer representation theorem for the integral kernel can be exploited \cite{Ferreira2009eigenvalues,CarmeliDeVito2006vector}, and there exists an optimal representation space to approximate the solution to this integral equation: the eigenspace the nonlinear integral operator $\mathcal{K}$.} While for the scaled dot-product attention's integral transform interpretation, explicitly pinning down the eigenspace is impossible during the dynamic training procedure. Nevertheless, we can still obtain a representation expansion to show that {the internal representations are propagated in a stable fashion} (Theorem \ref{theorem:stability}). 

\paragraph{Relation to Other Methods.} The integral transform formula derived in \eqref{eq:attention-kernel} resembles the nonlocal means method (NL-means) in tasks of traditional image signal processing such as denoising, \eg, \cite{BuadesCollEtAl2005review}. 
In NL-means, $\kappa(x,x')$ is a non-learnable integral kernel, which measures the similarity between pixels at $x$ and $x'$. While in ViT, the layer-wise kernel $\kappa(x,x'; \theta)$ is learnable, and establishes the inter-patch topology (continuity, total variation, \etc). 
One key common trait is that they are both normalized in the sense that for any $x\in \omega$
\begin{equation}
\label{eq:kernel-normalization}
    \alpha^{-1}(x)\int_{\omega} \kappa(x, x') \,d\mu (x')= 1,
\end{equation}
which is facilitated by the row-wise softmax operation to enforce a unit row sum. We shall see in Section \ref{sec:vit-stability} that this plays a key role in 
{obtaining} a stable representation in ViT layers mathematically.

Compared with another popular approach, the Convolutional Neural Network (CNN)-based models, the MAE has several differences. In CNN, the translation-invariance is inherited automatically from the convolution operation, \ie, the CNN kernel depends only on the difference of the positions $(x-x')$. 
Moreover, $K(\cdot)$ is acting on a pixel level and locally supported, thus having a small receptive field. For CNN to learn long-range dependencies, the deep stacking of convolutions is necessary but then renders the optimization algorithms difficult \cite{SzegedyLiuJiaSermanetEtAl2015Going}. 
Moreover, repeating local operations around a small pixel patch is computationally inefficient in that the convolution filter sizes being small in CNN makes the ``message passing'' from distant positions back and forth difficult \cite{WangGirshickEtAl2018Nonlocal}. {In MAE, the translation invariance is obtained through proper normalization. The learned kernel acts on the basis function representing each patch. Moreover, the kernel map is globally supported, which means it can learn effectively the interaction between even far away patches.} As a result, this global nature makes it easier to learn a better representation, thus greatly reduces the number of layers needed to complete the same task.






\section{Stable Representation Propagation in the Attention Block}
\label{sec:vit-stability}
In this section, we try to explain
why the representations are continuously changing in a stable fashion in both the encoder and decoder of MAE, which is first observed in \cite{RaghuUnterthinerEtAl2021ViT}. In the following theorem, we have proved a key result that the softmax normalization plays a vital role in stabilizing the propagation of the representations through the layers.

\begin{figure*}[!t]
	\centering
	\subfigure[]{
		\label{fig:input}
		\includegraphics[width=0.18\linewidth]{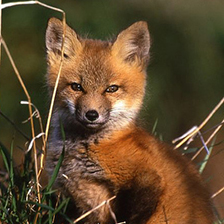}}
    \subfigure[]{
		\label{fig:output-enc}
		\includegraphics[width=0.18\linewidth]{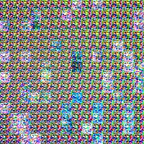}}
    \subfigure[]{
		\label{fig:latent-1}
		\includegraphics[width=0.18\linewidth]{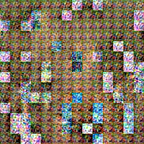}}
	\subfigure[]{
		\label{fig:latent-2}
		\includegraphics[width=0.18\linewidth]{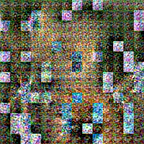}}
    \subfigure[]{
		\label{fig:latent-3}
		\includegraphics[width=0.18\linewidth]{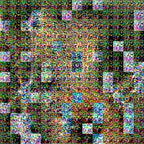}}
	\subfigure[]{
		\label{fig:latent-4}
		\includegraphics[width=0.18\linewidth]{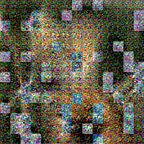}}
	\subfigure[]{
		\label{fig:latent-5}
		\includegraphics[width=0.18\linewidth]{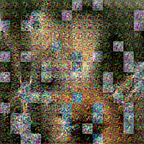}}
	\subfigure[]{
		\label{fig:latent-6}
		\includegraphics[width=0.18\linewidth]{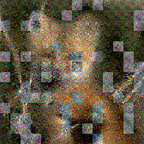}}
	\subfigure[]{
		\label{fig:latent-7}
		\includegraphics[width=0.18\linewidth]{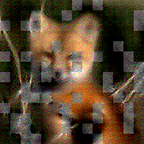}}
	\subfigure[]{
		\label{fig:output-dec}
		\includegraphics[width=0.18\linewidth]{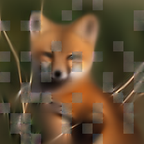}}
		
	\caption{The internal representation propagation during evaluation for a random sample from ImageNet-1K validation set. \subref{fig:input} input of MAE encoder; \subref{fig:output-enc} output of MAE encoder, concatenated with the learnable mask token embedding that is shared among all masked patches; \subref{fig:latent-1}--\subref{fig:latent-7} the latent representations, \ie, the outputs of the decoder layers $\#1$ to $\#7$; \subref{fig:output-dec} output of the last decoder layer ($\#8$). (We use the official released model and pretrained weights at
\url{https://github.com/facebookresearch/mae}.) See texts for details. Best viewed in color.}
	\label{fig:latent}
\end{figure*}

\begin{theorem}
\label{theorem:stability}
    {Assume that a trained attention block in the MAE/ViT encoder layer such that $ \|\mathbf{W}^Q- \mathbf{W}^K\|_2 $ are uniformly bounded above and below by two positive constants, and $\mathbf{W}^V=\mathbf{I}$, and the attention kernel being in the form of \eqref{eq:attention-normalized} such that it is translation invariant and symmetric, let $v^{(t)}(\cdot)$ and $v^{(t+1)}(\cdot)$ be the feature maps for the input and output of the scaled dot-product attention \eqref{eq:attention-expansion}, and $v^{(t)} \in BV(\omega) \cap \mathcal{V}$, then we have
\begin{equation}
    \|v^{(t+1)}  - v^{(t)}\|_{\mathcal{V}} \leq C(d) \frac{1}{n} \left(\|v^{(t)}\|_{\mathcal{V}} + |v^{(t)}|_{BV}\right) .
\end{equation}}
\end{theorem}
\paragraph{Proof Sketch and an Interpretation for Theorem \ref{theorem:stability}.}
The softmax operation in \eqref{eq:attention-dot-product} makes the attention matrix to have a row sum being 1. This normalization further translates to the integral kernel's integration being 1 in \eqref{eq:kernel-normalization}. This enables us to estimate of how the internal representations are propagated in a stable evolution from $v^{(t)}$ to $v^{(t+1)}$. We can rewrite $v^{(t)}(x)$
\begin{equation}
    v^{(t)}(x) = \alpha^{-1}(x)\int_{\omega} \kappa(x, x') v^{(t)}(x)  \,d\mu (x'),
\end{equation}

Thus, the propagation can lead the following modulus of continuity representation
\begin{equation}
\begin{aligned}
  &  v^{(t+1)}(x) - v^{(t)}(x)
\\
=&\; \alpha^{-1}(x)\int_{\omega} \kappa(x, x') v^{(t)}(x')  \,d\mu (x') - v^{(t)}(x)
\\
=&\; \alpha^{-1}(x)\int_{\omega} \kappa(x, x') \bigl(v^{(t)}(x') - v^{(t)}(x)\bigr)  \,d\mu (x').
\end{aligned}  
\end{equation}
Using the argument in Section \ref{paragraph:rkhs}, $\kappa(x, x')=K(x-x')$ thus a simple substitution can be exploited to get the following representation:
\begin{equation}
    \begin{aligned}
& \alpha^{-1}(x)\int_{\omega} K(x-x') \bigl(v^{(t)}(x') - v^{(t)}(x)\bigr)  \,d\mu (x')
\\
= &\; \alpha^{-1}(x)\int_{\omega_{\xi}} K(\xi) \bigl(v^{(t)}(x+\xi) - v^{(t)}(x)\bigr)  \,d\mu( \xi).
\end{aligned}  
\end{equation}
By Mercer's theorem, there exists an eigenfunction expansion for $\kappa(x,x')= \sum_{i=1}^{\infty} a_i\psi_i(x)\psi_i(x')$ that has a spectral decay. Then, the propagation bounds can be proved {under the assumption that the current layer's learned feature map offers a ``reasonable'' approximation to the eigenspace of $\kappa(\cdot,\cdot)$}. 




Overall, this bound describes that the propagation of the internal representation is continuously changing based on the inter-patch interaction encoded in the attention kernel. 
In the kernel interpretation of the attention mechanism, there is one key difference with the conventional kernel method: {the conventional kernel measures the inter-sample similarity in the whole dataset; while the attention kernel $\kappa(\cdot,\cdot)$ or $\tilde{\kappa}(\cdot,\cdot)$ is built for a single instance, and learns inter-patch topological relations to build a better representation space for the MAE.}
{
This learned representation for a specific data sample determines how amenable it is for the downstream tasks.}

Additionally, we note that this result applies to the layer-wise representation propagation in the decoder layers as well. An enlightening illustration is demonstrated in Figure \ref{fig:latent}. {There are multi-fold interesting aspects about this evolution diagram shown}: (1) in the ViT-base MAE, the embedding dimension of its decoder is merely $512$, which is less than the embedding dimension in the encoder ($d=768$). Note that $768=N_c\times N_c\times 3$ which enables that the latent representation (a row vector) in a single patch can be directly reshaped to an image patch. Thus, we need a patch-wise upsampling projection; (2) this projection, which maps vectors in $\mathbb{R}^{196\times 512}$ to those in $\mathbb{R}^{196\times 768}$ resides in the MAE decoder. It should be noted that this upsampling projection is only connected with the last decoder layer. Heuristically, this projection {may} only upsample the last decoder layer's output $\mathbf{z}_8\in \mathbb{R}^{196\times 512}$ to the desired embedding dimensions. Nevertheless, due to the stable representation propagation, the outputs from the previous decoder layers can also benefit from the last layer projection weights to obtain sensible reconstruction results, as illustrated in Figure \ref{fig:latent}.

\paragraph{Skip-Connection as a Tikhonov Regularizer} 
After presenting Theorem \ref{theorem:stability}, one might ask: given the integral transform representation is already stable, what is the role of the skip-connection? Diverting from the original interpretation of battling the diminishing gradient for the skip-connection 
\cite{HeZhangEtAl2016Deep}, here we offer a new perspective and some heuristics inspired by functional analysis to articulate the reason why using the skip-connection would make the representation propagation more stable, which is also observed in \cite{RaghuUnterthinerEtAl2021ViT}.

Knowing that the integral kernel interpretation of attention \eqref{eq:attention-kernel} resembles the Fredholm integral equation of the first kind, we can interpret the skip-connection as a layer-wise Tikhonov regularization in the Fredholm integral equation of the second kind.
Starting from \eqref{eq:attention-kernel}
\begin{equation}
z(x) =  \alpha^{-1} \int_{\omega}  \kappa(x,x') v(x') d\mu (x').
\end{equation}
This Fredholm intergral equation of the first kind is usually extremely ill-posed \cite{Hadamard2003Lectures}, in the sense that the solution, even if it exists, does
not depend continuously on $v(\cdot)$. 

If $\mathbf{W}^V \approx \beta^{-1} \mathbf{I}$, then for $x\in \mathcal{X}$, we have that 
\begin{equation}
\label{eq:fredholm}
z(x) \approx \beta v(x) + \alpha^{-1}  \int_{\omega} \kappa(x,x') v(x') d\mu (x').
\end{equation}
This is the Fredholm integral equation of the second kind witha variable coefficient, the extra $\beta v(\cdot)$ term not only contributes to the well-posedness of this equation, but renders the numerical scheme to approximate a better $v(\cdot)$ more stable. Tikhonov firstly introduced this method in \cite{Tihonov1963Solution}, see also \cite[Chapter 16]{Kress2014Linear}.

\begin{theorem}[Skip-connection can represent the minimizer to a Tikhonov-regularized integral equation functional]
\label{theorem:tikhonov}
For $\mathcal{K}(\cdot)$ using $\kappa(\cdot,\cdot)$ as the integral kernel in \eqref{eq:attention-normalized}, define the following functional 
\begin{equation}
\label{eq:tikhonov}
    \mathcal{L}:= \frac{1}{2}\|\alpha^{-1}(\cdot) \mathcal{K}(v) - z\|_{\mathcal{V}*}^2 + 
  \beta\langle \mathcal{K}(v), v\rangle,
\end{equation}
where $\|u\|_{\mathcal{V}'} := \sup_{v\in \mathcal{V}} |\langle u, v\rangle|/\|v\|_{\mathcal{V}}$ denotes the dual norm. 
The Euler-Lagrange equation associated with $\min_{v}\mathcal{L}$ has the following form:
\begin{equation}
\label{eq:tikhonov-eq}
   \beta u + \alpha^{-1}\mathcal{K}(u) = z \quad \text{ in } \; (\mathcal{K}^*(\mathcal{V}))'.
\end{equation}
\end{theorem}
Note that \eqref{eq:tikhonov-eq} bears exactly the same form with \eqref{eq:fredholm}. Moreover, instead of the conventional $L^2$-type Tikhonov regularizer $\|v\|_{\mathcal{V}}^2$, here $\langle \mathcal{K}(v), v\rangle$ is an induced norm by the positive (semi)definite integral operator $\mathcal{K}(\cdot)$.
The skip-connection term in \eqref{eq:tikhonov-eq} comes from the Tikhonov regularization term in \eqref{eq:tikhonov}, and without it, the Euler-Lagrange equation reverts to the Fredholm integral equation of the first kind.



%% file: 3_mae.tex

\section{MAE Decoder: Low-Rank Reconstruction Through Global Interpolation}
\label{sec:MAE-decoder}

During pretraining, the major function of the MAE decoder is to map the low-rank representation obtained by the MAE encoder to a reconstruction.
The encoder embedding has a bigger dimension, yet is define on only a fraction of patches. The decoder reduces the embedding dimension, but is able to obtain the embedding for all $p\times p$ patches including the masked ones. 
Despite of the fact that the decoder in MAE is only used in pretraining, not downstream tasks, it plays a vital role in learning a ``good'' representation space for the MAE encoder.





\paragraph{Enrichment of the Representation Space through Positional Embedding}
{Because the positional embedding $\mathbf{x}:=\Vert_{i=1}^n x_i \in \mathbb{R}^{p\times d}$ is added to the latent representation $\mathbf{y}$ in \eqref{eq:attention}}, the nonlinear universal approximator (FFN) in each attention block shall also contribute to learning a better representation.
In every decoder layer, the basis functions in $\mathcal{V}$ are being constantly enriched by 
$\operatorname{span}\{w_j\in \mathcal{X}: w_j(x_i) = (\text{FFN}(\mathbf{y}+\mathbf{x}))_{ij}, 1\leq j\leq d \}$. If we treat the positional embeddings $\{x_i\}$ as coordinate again, $\text{FFN}(\mathbf{y}+\mathbf{x})$ is a subset of the famous Barron space \cite{Barron1993Universal,Bach2017Breaking}, which has a rich and 
{powerful representation power that can approximate smooth function arbitrarily well}, \eg, see \cite{HutzenthalerJentzenEtAl2020proof,Xu2020finite}. As a result, the representation space is dynamically updated layer by layer to try to build a more expressive representation to better characterize the inter-patch topology. 
FFNs themselves have no linear structure, however, the basis functions produced this way act as a building block to update the linear space for the expansion in \eqref{eq:attention-expansion} and \eqref{eq:attention-kernel}. 
In the MAE decoder, the number of basis ($p^2=196$) is much greater than that of the MAE encoder ($0.25 \cdot p^2 = 49$), thus heuristically speaking, this basis update mechanism is mostly working in the decoder.


\paragraph{Reconstruction is a Global Interpolation.}

In MAE decoder, the mask token, a learnable vector shared by every masked patch, is concatenated to the unshuffled representation of the unmasked patches. To demonstrate that the reconstructed image is a global interpolation by using the unmasked patch embeddings, we first need to prove the following lemma, which states that the learned mask token embedding can be simply replaced by 
{zero with an updated set of attention weights.} 
For the simplicity, we denote 
the embedding dimension in the MAE decoder still as $d$. 

\begin{lemma}
Let $\mathbf{m}\in \mathbb{R}^{1\times d}$ be the learned mask token embedding that is shared by all masked patches, and let $m(\cdot)$ be its feature map.
Denote the set of masked and unmasked patches' index as $\mathcal{M}$ and $\mathcal{N}$, respectively, and we assume that $|\mathcal{N}|\geq 2$. For the input $\mathbf{y}\in \mathbb{R}^{p\times d}$ that already contains the masked patches, i.e., all $i\in \mathcal{M}$, $\mathbf{y}_i=\mathbf{m}$,
, there exists a new set of affine linear maps to generate $\{\widehat{\mathbf{Q}}, \widehat{\mathbf{K}}, \widehat{\mathbf{V}}\}$, such that for any $i=1,\dots, n$,
\begin{equation}
\left\| \sum_{j=1}^{n} A(\mathbf{q}_i, \mathbf{k}_j) \, \mathbf{v}_i
- \sum_{j\in \mathcal{N}} A(\hat{\mathbf{q}}_i, \hat{\mathbf{k}}_j) \, \hat{\mathbf{v}}_i\right\|
< Cn^{-1}.
\end{equation}
\end{lemma}

With this lemma, we are already present our final result: for the first layer of the MAE decoder, the network interpolates the representation using global information from the embeddings learned by the MAE encoder, not just the ones from the nearby patches. Moreover, with Theorem \ref{theorem:stability}, the MAE decoders continue to {perform such a global interpolation in subsequent layers}. For an empirical evidence, please refer to Figure \ref{fig:latent}.

\begin{proposition}[{Interpolation results for masked patches}]
{For the embedding of every masked patch $i\in \mathcal{M}$,  let $v_i^{(t+1)}$ be the output embedding of a decoder layer for this patch, and let $v_j^{(t)}$ be the input from the encoder for $j=1,\dots, p$}, then $v_i^{(t+1)}$ is 
\begin{equation}
  v_i^{(t+1)} = \sum_{j\in \mathcal{N}} a_j v_j^{(t)},
\end{equation}
for a set of weights based on unmasked patches $a_j(v_{i_1}, \dots, v_{i_k})$, $\mathcal{N} := \{i_1, \dots, i_k\}$. Moreover, the reconstruction quality is bounded above by the global reconstruction error of the unmasked patches,
\begin{equation}
  \|R v_i - \mathbf{u}_i\|_{BV(\Omega_i)} \leq C \sup_{j\in \mathcal{N}}\|R v_j - \mathbf{u}\|_{BV(\Omega_j)} + C n^{-1} \|\mathbf{u}\|.
\end{equation}
\end{proposition}




%% file: 4_appendix.tex

\appendix



\section{Proof of Theorem \ref{theorem:stability}}

\begin{assumption}[Assumption of Theorem \ref{theorem:stability}]
\label{assumption:stability}
To prove Theorem \ref{theorem:stability-rigor}, we assume the following conditions hold true,
\begin{enumerate}[align=left, leftmargin=10pt,
                  label=$\emph{(}$\subscript{B}{{\arabic*}}$\emph{)}$\;]
  \item \label{asp:regularity} The underlying Hilbert space for the latent representation is $\mathcal{V} \subset \big(L^2(\omega, \mu))^d$, i.e., there is no \emph{a priori} inter-channel continuity, and the inter-channel topology is learned from data.   

  \item \label{asp:extension} For a BV function on $\omega$, there exists a smooth extension to $\mathbb{R}^2$ such that the extension is stable in $|\cdot|_{BV(\Omega)}$ (see \eg, \cite{Gerhardt2006Analysis},
\cite[Theorem 2]{BuragoMazya1969}).
  \item For each $v\in \mathcal{V}$, there exists a smooth $K$ such that $\kappa(x,x')=K(x-x')$, and $K(x-x')\leq C_{\gamma}e^{ -\gamma \|x-x'\| }$ with a uniform constant $C_\gamma$ that only depends on $\gamma$.
  \item For each $v\in \mathcal{V}$, $ \alpha(x) := \int_{\omega} \kappa(x, x') \,d\mu (x')$, and $\alpha$ is bounded below by a positive constant $C_{\alpha}$ uniformly.
    %
  
\end{enumerate}
\end{assumption}

\begin{manualtheorem}{5.1}[Stability result, rigorous version]
\label{theorem:stability-rigor}
Under the assumption of Assumption \ref{assumption:stability}, we have
\begin{equation}
    \|v^{(t+1)}  - v^{(t)}\|_{\mathcal{V}} \leq C(\gamma,d) \frac{1}{n} \left(\|v^{(t)}\|_{\mathcal{V}} + |v^{(t)}|_{BV}\right) .
\end{equation}
\end{manualtheorem}

\begin{proof}
First we extend the kernel function from $\omega$ to the whole $\mathbb{R}^2$ by 
\begin{equation}
    \operatorname{Ext}(\kappa(x,x')) = 
\begin{cases}
    \kappa(x,x') & \text{if } x \text{ and } x' \in \omega,
\\
    0 & \text{otherwise}.
\end{cases}
\end{equation}
With a slight abuse of notation we still denote the extension $\operatorname{Ext}(\kappa(\cdot,\cdot))$ still as 
$\kappa(\cdot,\cdot)$, and the normalization \eqref{eq:kernel-normalization} still holds:
\begin{equation}
\label{eq:kernel-normalization-R2}
    \alpha^{-1}(x)\int_{\omega}  \kappa(x, x') \,d\mu (x') =  \alpha^{-1}(x)\int_{\mathbb{R}^2}  \kappa(x, x') \,d\mu (x') = 1.
\end{equation}
Without loss of generality, we also assume that the 
Thus, under assumption \hyperref[asp:regularity]{$(B_{1})$}--\hyperref[asp:extension]{$(B_{2})$}, we have the original integral on $\omega$ can be written as on the full plane
\begin{equation}
    v^{(t+1)}(x) = \alpha^{-1}(x)\int_{\mathbb{R}^2} \kappa(x, x') v^{(t)}(x')  \,d\mu (x').
\end{equation}
Similarly by \eqref{eq:kernel-normalization-R2}, 
\begin{equation}
    v^{(t)}(x) = \alpha^{-1}(x)\int_{\mathbb{R}^2} \kappa(x, x') v^{(t)}(x)  \,d\mu (x').
\end{equation}
Therefore,

\begin{equation}
\begin{aligned}
  &  v^{(t+1)}(x) - v^{(t)}(x)
\\
=&\; \alpha^{-1}(x)\int_{\mathbb{R}^2} \kappa(x, x') \bigl(v^{(t)}(x') - v^{(t)}(x)\bigr)  \,d\mu (x').
\end{aligned}  
\end{equation}

Using $\kappa(x, x')=K(x-x')$, let $\xi = x'-x$:
\begin{equation}
    \begin{aligned}
& \alpha^{-1}(x)\int_{\mathbb{R}^2} K(x-x') \bigl(v^{(t)}(x') - v^{(t)}(x)\bigr)  \,d\mu (x')
\\
= &\; \alpha^{-1}(x)\int_{\mathbb{R}^2} K(\xi) \bigl(v^{(t)}(x+\xi) - v^{(t)}(x)\bigr)  \,d\mu( \xi).
\end{aligned}  
\end{equation}

By $v^{(t)} \in BV(\mathbb{R}^2)$, and $\omega$ being compact, there exists $\delta>0$ such that 
\begin{equation}
    \sup_{x\in \omega}\sup_{\|\xi\|< \delta/n} \|v^{(t)}(x+\xi) - v^{(t)}(x) \| < \frac{1}{n} |v^{t}|_{BV(\mathbb{R}^2)}.
\end{equation}

Thus, for any ${x\in \omega}$
\begin{equation}
\begin{aligned}
    &  \| v^{(t+1)}(x) - v^{(t)}(x)\|
\\
\leq &\; |\alpha^{-1}| \int_{\mathbb{R}^2} \| K(\xi) \bigl(v^{(t)}(x+\xi) - v^{(t)}(x)\bigr) \| \,d\mu( \xi)
\\
  \leq &\;  |\alpha^{-1}| \sup_{\|\xi\|< \delta/n}\|v^{(t)}(x+\xi) - v^{(t)}(x)\| \int_{\omega}\left|K(\xi)\right| d\mu( \xi)
\\
 & +|\alpha^{-1}|\int_{\|\xi\|\geq \delta/n}\left|K(\xi)\right| 2\|v^{(t)}\|_{\infty} d \mu( \xi)
\\
\leq &\; 2C_{\alpha}^{-1} C_{\gamma} \frac{1}{n} 
\left(|v^{(t)}|_{\infty} + |v^{(t)}|_{BV} \right).
\end{aligned}
\end{equation}
Taking $\sup_{x\in \omega}$ yields the result.
\end{proof}

\section{Proof of Theorem \ref{theorem:tikhonov}}
\begin{proof}
    
Define $\eta(\cdot): \mathcal{V}\to \mathbb{R}$ such that 
$\eta_u(v) := \langle z, v\rangle -  \langle \alpha^{-1} \mathcal{K}u, v\rangle$, where $u\in \mathcal{W}\subseteq \mathcal{V}$ is the solution to the integral equation in \eqref{eq:tikhonov-eq}, and $\mathcal{W}$ is the solution subspace.
Clearly,
$\eta_u\in \mathcal{V}'$ defines a bounded functional on $\mathcal{V}$ for $(\mathcal{V}, \langle\cdot, \cdot\rangle)$.  By Riesz representation theorem, there exists an isomorphism $G: \mathcal{V} \to \mathcal{V}'$ such that $\phi_u := G^{-1}(\eta)\in \mathcal{V}$
and
\begin{equation}
\label{eq:mix-1}  
\eta_u(v) = \langle \phi_u, v\rangle =  \langle z, v\rangle - \langle \alpha^{-1}\mathcal{K}(u), v\rangle.
\end{equation}
Then, define $\|\cdot\|_{\mathcal{V}}^2 := \langle\cdot, \cdot\rangle$, we have
\[
\min_{u\in \mathcal{V}} \|\eta(\cdot)\|_{\mathcal{V}'}^2
= \min_{u\in \mathcal{V}} \|\phi_u\|_{\mathcal{V}}^2
= 
\min_{u\in \mathcal{V}} \|G^{-1}(\langle z, \cdot\rangle -  \langle\alpha^{-1}(\cdot) \mathcal{K}(u), \cdot\rangle )\|_{\mathcal{V}}^2.
\]
Thus, the functional $\mathcal{L}(v)$ can be written as:
\begin{equation}
   \mathcal{L}(v) =  \frac{1}{2}\left\{\|\alpha^{-1}(\cdot) \mathcal{K}(v) - z\|_{\mathcal{V}'}^2 + 
  \beta\langle \mathcal{K}(v), v\rangle\right\}
\end{equation}

Taking the Gateaux derivative $\lim_{\tau\to 0} \dd \mathcal{L}(u+\tau w)/ \dd \tau$ in order to find the critical point(s) $u\in \mathcal{W}$, we have for any perturbation $w\in \mathcal{W}$ such that $u+\tau w \in \mathcal{W}$
\[
\begin{aligned}
    0 = \lim_{\tau\to 0}\frac{d}{d\tau} 
\Bigg\{
& \left\langle\alpha^{-1}(\cdot) 
G^{-1}(\langle z, \cdot\rangle - \alpha^{-1}(\cdot) \mathcal{K}(u+\tau v), \cdot\rangle ),
G^{-1}(\langle z, \cdot\rangle - \alpha^{-1}(\cdot) \mathcal{K}(u+\tau v), \cdot\rangle )
\right\rangle ,
\\ &+ \beta\langle \mathcal{K}(u+\tau v), u+\tau v\rangle 
\Bigg\}
\end{aligned}
\]
and applying $G^{-1}$ on $\mathcal{K}(\cdot, u)\in \mathcal{V}'$, it reads for any $w\in \mathcal{W}$
\begin{equation}
\label{eq:mix-2}
\begin{aligned}
& \left\langle
 \alpha^{-1}(\cdot) G^{-1}\big(\langle z, \cdot\rangle  - \langle \mathcal{K}(u),\cdot\rangle \big),
G^{-1}(\langle \mathcal{K}(w),\cdot\rangle) \right\rangle + \langle \mathcal{K}(u), v \rangle
\\
=&\; \left\langle \alpha^{-1}  \phi_u, G^{-1}(\langle \mathcal{K}(u),\cdot\rangle )\right\rangle
+ \langle \mathcal{K}(u),w \rangle
\\
= &\; \langle \alpha^{-1}\mathcal{K}(w),\phi_u\rangle + \langle \mathcal{K}(u),w\rangle
= 0.   
\end{aligned}
\end{equation}
As a result, combining \eqref{eq:mix-2}, \eqref{eq:mix-1}, and the self-adjointness of $\mathcal{K}$, we have the system in the continuum becomes
\begin{equation}
\label{eq:mix}
\left\{
\begin{aligned}
 \langle \phi_u, v\rangle + \langle\alpha^{-1}\mathcal{K}(u), v \rangle  & = \langle z, v\rangle, & \forall v\in \mathcal{V},
\\
 \langle \alpha^{-1}\mathcal{K}(w), \phi_u \rangle + \beta\langle \mathcal{K}(u), w\rangle &= 0, & \forall q \in \mathcal{Q}.
\end{aligned}
\right.
\end{equation}
The first equation implies that: 
\begin{equation}
    \phi_u + \alpha^{-1}\mathcal{K}(u) = z \quad \text{ in }\; \mathcal{V}'.
\end{equation}
Hence, by $w\in \mathcal{W}\subseteq \mathcal{V}$, when plugging $w$ to the second equation in \eqref{eq:mix-1} we have 
\begin{equation}
    \langle z - \alpha^{-1} \mathcal{K}(u) + \beta u, \mathcal{K}(w)\rangle = 0 \quad \text{ for } w\in \mathcal{W},
\end{equation}
and the theorem follows.

\end{proof}